\documentclass[sigconf,natbib=false]{acmart}

\AtBeginDocument{%
  }

\setcopyright{acmcopyright}
\setcopyright{none}
\acmYear{2008}
\acmDOI{10.1145/1463434.1463464}

\acmConference[ACM GIS '08]{16th ACM SIGSPATIAL International Conference on Advances in Geographic Information Systems}{June 5--7, 2008}{Irvine, CA, USA.}
\acmPrice{5.00}
\acmISBN{978-1-60558-323-5/08/11}

\RequirePackage[
  datamodel=acmdatamodel,
  style=acmnumeric,
  ]{biblatex}

\begin{document}

\title{Validation of Vector Data using Oblique Images}


\author{Pragyana Mishra}
\affiliation{%
 \institution{Microsoft Corporation}
 \streetaddress{One Microsoft Way}
 \city{Redmond}
 \state{Washington}
 \country{USA}}
\email{pragyanm@microsoft.com}

\author{Eyal Ofek}
\affiliation{%
 \institution{Microsoft Corporation}
 \streetaddress{One Microsoft Way}
 \city{Redmond}
 \state{Washington}
 \country{USA}}
\email{eyalofek@microsoft.com}

\author{Gur Kimchi}
\affiliation{%
 \institution{Microsoft Corporation}
 \streetaddress{One Microsoft Way}
 \city{Redmond}
 \state{Washington}
 \country{USA}}
\email{gkimchi@microsoft.com}

\renewcommand{\shortauthors}{Mishra et al.}

\newcommand{\vect}[1]{\boldsymbol{#1}}

\setlength{\belowcaptionskip}{-10pt}

\begin{abstract}
Oblique images are aerial photographs taken at oblique angles to the earth’s surface. Projections of vector and other geospatial data in these images depend on camera parameters, positions of the geospatial entities, surface terrain, occlusions, and visibility. This paper presents a robust and scalable algorithm to detect inconsistencies in vector data using oblique images. The algorithm uses image descriptors to encode the local appearance of a geospatial entity in images. These image descriptors combine color, pixel-intensity gradients, texture, and steerable filter responses. A Support Vector Machine classifier is trained to detect image descriptors that are not consistent with underlying vector data, digital elevation maps, building models, and camera parameters. In this paper, we train the classifier on visible road segments and non-road data. Thereafter, the trained classifier detects inconsistencies in vectors, which include both occluded and misaligned road segments. The consistent road segments validate our vector, DEM, and 3-D model data for those areas while inconsistent segments point out errors. We further show that a search for descriptors that are consistent with visible road segments in the neighborhood of a misaligned road yields the desired road alignment that is consistent with pixels in the image.
\end{abstract}



\maketitle

\section{Introduction}
Generating a map may involve several sources of geospatial information --- a few of which include vector data, map projection parameters, digital elevation maps (DEM’s), and 3-dimensional (3-D) models. Errors or inaccuracies in these data sources contribute to the overall accuracy and quality of the map. Detecting these errors is difficult as there is little to no ground truth available against which the data can be compared and validated. 

Most aerial and satellite images that may be used as ground truth for data validation are nadir views of the world. However, the set of attributes of data that can be validated using aerial images is limited. This is due to the fact that the projection is orthographic in nature and 3-D attributes of the scene---from both terrain and building structures---are not entirely captured in nadir, orthographic views. 

Inaccuracies in altitude, heights of buildings, and vertical positions of structures cannot be easily detected by means of nadir imagery. These inaccuracies can be detected in an image where the camera views the scene at an oblique angle. Oblique-viewing angles enable registration of altitude-dependent features. Additionally, oblique viewing angles bring out the view-dependency of 3-D geospatial data; occluding and occluded parts of the world are viewed in the image depending on the position and orientation (pose) at which the image was taken. Image pixel regions that correspond to occluding and occluded objects can be used to derive consistency of the geospatial entities as they appear in the image. The consistency can be further used to rectify any inaccuracies in data used for segmenting and labeling of the objects. 

Oblique images are aerial photographs of a scene taken at oblique angles to the earth’s surface. These images have a wide field-of-view and therefore, cover a large expanse of the scene, as can be seen in Figures \ref{Figure04} through \ref{Figure07}. Oblique images capture the terrain of earth’s surface and geometrical structures such as buildings and highways. For example, straight roads that are on the slope of a hill will appear curved. The mapping between a 3-D point of the world and its corresponding point on the image is a non-linear function that depends on the elevation, latitude, and longitude of the 3-D point in space. Additionally, visibility of the 3-D point depends on the camera’s position and orientation, and the structure of the world in the neighborhood of that 3-D point; the point may not be visible in the image if it is occluded by a physical structure in the line-of-sight of the camera (as illustrated in Figure \ref{Figure01}. These properties can, therefore, be used to detect and rectify any errors of the 3-D point’s position as well as any inconsistencies of structures in the neighborhood of the point.

We show that pixels of oblique images can be analyzed at scale to detect, validate, and correct disparate sources of data, such as vectors, projection parameters of camera, digital elevation maps (DEM’s), and 3-D models. This is a novel use of oblique imagery wherein errors due to inaccurate and imprecise vectors, DEM, map projection parameters, and 3-D model data are detected and rectified using pixel statistics of oblique images. We present a statistical-learning algorithm to learn image-based descriptors that encode visible data consistencies, and then apply the learnt descriptors to classify errors and inconsistencies in geospatial data. The algorithm combines different descriptors such as color, texture, image-gradients, and filter responses to robustly detect, validate, and correct inconsistencies.

\section{Oblique Images}

Oblique images capture the view-dependency and height of geospatial data. This unique characteristic makes oblique images ideal for validation and correction of disparate data sources. The projection of the 3-D world accounts for altitude (terrain), structure models, and the arrangement of geospatial entities with respect to each other. Both the horizontal and vertical attributes, which are visible in oblique views, can be used to constrain the set of possible hypotheses for generating attributes of geospatial data. Vertical attributions such as visible building facets can be used to validate geospatial structures and models. The advantages of using oblique views over nadir views are also discussed in the research on generating 3-D building models from monocular aerial images \cite{ProjectiveObject1994, PerformanceMonocularBuilding1999}. 

The oblique angle, i.e., the angle formed by the optical axis of the camera with respect to the $Z$-axis of the world coordinate system, of our oblique images varies between 30 to 45 degrees. This falls in the category of high-oblique photography. In photogrammetry, a nadir or vertically-viewed image has a small tilt angle, typically on the order of five degrees; a low-oblique image is considered to be one that has between 5 and 30 degrees of tilt, and high-oblique image has greater than 30 degrees of tilt \cite{PerformanceMonocularBuilding1999}. The term \emph{oblique image} refers to a high-oblique image in this paper. In order to understand and learn appearances of 3-D geospatial entities as observed in images, image-pixel statistics are first computed using a set of image-based descriptors. A Support Vector Machine (SVM) classifier is then trained for learning the descriptors. We have chosen a SVM classifier because of its simplicity and our ability to analyze and explain any inconsistencies in the vector data as delineated by the classifier. The classifier validates points that are consistent with learnt descriptors for a given geospatial entity and detects those that are not. In this paper, the chosen geospatial entity is road data. However, our classifier can be trained to detect consistency of other geospatial entities such as buildings, water bodies, foliage, parking lots, bridges, and railway lines.

\section{Image Descriptors}

Image descriptors encapsulate pixel statistics of local image regions. These descriptors encapsulate the characteristic appearance of a geospatial entity as it appears in an image. The choice of descriptors primarily depends on two criteria: first, the descriptors of a geospatial entity can be learnt by a classifier, and second, a descriptor that belongs to the imaged geospatial entity can be separated from a descriptor of an image region that does not correspond to that geospatial entity.

\begin{figure}[ht]
  \centering
  \includegraphics[width=\linewidth]{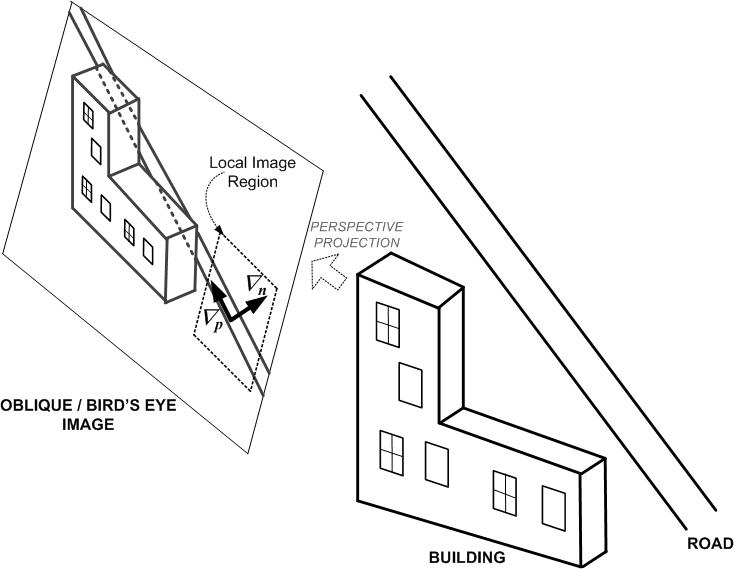}
  \caption{The local image region, and the primary and normal gradient directions for the projected road segment in the image plane. Note that the building may occlude the view of the road in the oblique or bird's eye image.}
  \Description{Image gradients along projected road vectors}
  \label{Figure01}
\end{figure}

Oblique images have certain amount of noise and clutter. Therefore, each image is pre-processed by smoothing the image with a Gaussian kernel of standard deviation $\sigma_s$. This smoothing kernel is applied to image regions along each road segment instead of the whole image for computational efficiency. The smoothing in our case was set to $\sigma_s$ = 2.8 pixels for the given image dimensions. The descriptor captures both the visual shape and the color characteristics of the local image region. The smoothed pixel region is bias and gain normalized in each color channel. The bias-gain normalized color value is computed by subtracting the mean and dividing by the standard deviation of the region. The following sections discuss how the local descriptors are computed by considering pixel regions around projected road-vector segments.

\subsection{Color Moments}
Color is a very important descriptor in extracting information specific to a geospatial entity from images. In our algorithm, color of a region is represented by local histograms of photometric invariants. This is derived from the pixel values in $CIELUV$ space. From all the pixel values in the region, the three central moments---median, standard deviation, and skewness---for each color component are computed and concatenated into a 9-dimesional color vector. Each dimension of this vector is normalized to unit variance so that the $L$ (luminance) component does not dominate the distance metric. The $CIELUV$ color space \cite{ColorScience1982} has perceptual uniformity---which means that any pair of equidistant points in the color space has the same perceived color difference. Other color spaces such as $RGB$, $HSV$, $CIE$, and $XYZ$ do not exhibit perceptual uniformity. 

Color moments for each component of the $CIELUV$ color space result in a 9-dimensional color descriptor---three moments for each of the three components. Color moments have proven to be consistent and robust for encoding color of image patches \cite{BenchmarkingImageFeatures1998}. The idea of using color moments to represent color distributions was originally proposed in \cite{SimilarityColorImages1995}. It is a compact representation and it has been shown in \cite{BenchmarkingImageFeatures1998} that the performance of color moments is only slightly worse than high-dimensional color histograms. However, their accuracy is consistently better than cumulative color histograms \cite{SimilarityColorImages1995}. However, color moments lack information about how the color is distributed spatially.

\subsection{Normalized Gradient}
The gradient vector of each bias-gain normalized region is evaluated along the primary and normal direction: $\nabla_p$ and $\nabla_n$ as illustrated in Figure \ref{Figure01}. The evaluated gradient is transformed into a positive-valued vector of length 4 whose elements are the positive and negative components of the two gradients. The vector, [ $|\nabla_p |-\nabla_p$, $|\nabla_p |+\nabla_p$, $|\nabla_n | -\nabla_n$, $|\nabla_n | +\nabla_n$ ], is a quantization of orientation in the four directions. Normalized gradients and multi-orientation filter banks, such as steerable filters, when combined with the color descriptor encode the appearance of the geospatial entity in a local neighborhood instead of just sampling pixel statistics at a point or a few points.

\subsection{Steerable Filter Response}
Steerable filters \cite{SteerableFilters1991} are applied at each sampled position along a vector-road segment along the primary and normal directions. The magnitude response from quadrature pairs using these two orientations on the image plane yields two values. A vector of length 8 can be further computed from the rectified quadrature pair responses directly. The steerable filters are based on second-derivative. We normalize all the descriptors to unit length after projection onto the descriptor subspace. The pixel regions used in computing the descriptors were patches of 24 x 24 pixels sampled at every 12 pixel distance along the projected road segments.

\subsection{Eigenvalues of Hessian}
The Hessian matrix comprises the second partial derivatives of pixel intensities in the region. The Hessian is a real-valued and symmetric 2 x 2 matrix. The two eigenvalues are used for the classification vector. Two small eigenvalues mean approximately uniform intensity profile within the region. Two large eigenvalues may represent corners, salt-and-pepper textures, or any other non-directional texture. A large and a small eigenvalue, on the other hand, correspond to a unidirectional texture pattern.

\subsection{Difference of Gaussians}
The smoothed patch is convolved with another two Gaussians---the first $\sigma_{s1}$ = 1.6 $\sigma_s$ and the second $\sigma_{s2}$ = 1.6 $\sigma_{s1}$. The pre-smoothed region, $\sigma_s$, is used as the first Difference-of-Gaussians (DoG) center and the wider Gaussian of resolution $\sigma_{s1}$ = 1.6 $\sigma_s$ is used as the second DoG center. The difference between the Gaussians pairs---that is the difference between $\sigma_{s1}$ and $\sigma_s$ and the difference between $\sigma_{s2}$ and $\sigma_{s1}$)---produces two linear DoG filter responses. The positive and negative rectified components of these two responses yield a 4-dimensional descriptor.

\section{Descriptor Classifier}
While validating vector data, road vectors are classified into two classes---the first class being visible and consistent road segments while the second class consists of inconsistent segments. This classification uses training data with binary labels. We have used a Support Vector Machine (SVM) with a non-linear kernel for this binary classification. In their simplest form, SVMs are hyperplanes that separate the training data by a maximal margin \cite{LearningWithKernels2018}. The training descriptors that lie on closest to the hyperplane are called support vectors. We briefly discuss how a SVM classifier works.

Consider a set of $N$ linearly-separable training samples of descriptors $\vect{d_i} \: (i = 1 \ldots \; N)$, which are associated with the classification label $\vect{c_i} \: (i = 1 \ldots \; N)$. The descriptors in our case are composed of the vector of color moments, normalized gradients, steerable filter responses, eigenvalues of Hessian, and difference of Gaussians. The binary classification label, $c_i \in \{ -1, \; +1 \}$, is $\: +1$ for sampled road vectors that are consistent with underlying image regions and $-1$ for points where road segments are inconsistent with corresponding image regions. This inconsistency, as we have pointed out earlier, could be due to inaccurate vector, DEM, building model data, or camera projection parameters.

The set of hyperplanes separating the two classes of descriptors is of the general form $\langle \vect{w}, \vect{d} \rangle \: + \: b \: = \: 0$.  $\langle \vect{w}, \vect{d} \rangle$ is the dot product between a weight vector $\vect{w}$ and $\vect{d}$, while $b \in \mathbb{R}$ is the bias term. 
Re-scaling $\: \vect{w}$ and $b \:$ such that the point closest to the hyperplane satisfy $|\: \langle \vect{w}, \vect{d} \rangle + b \:| = 1$, we have a canonical form ($\vect{w}$, $b$) of the hyperplane that satisfies $c_i (\langle \vect{w}, \vect{d} \rangle + b) \geq 1$ for $i = 1 \ldots \; N$. For this form the distance of the closest descriptor to the hyperplane is 1/||$\vect{w}$||. For the optimal hyperplane this distance has to be maximized, that is:
\begin{displaymath}
  \mathrm{minimize} \; \; \varphi (\vect{w}) =  \frac{1}{2} || \vect{w}||^2 
\end{displaymath}
\hspace{3em} subject to $c_i (\langle \vect{w}, \vect{d} \rangle + b) \geq 1$ for all $\: i = 1 \ldots \; N$

\smallskip \noindent
The objective function $\varphi( \vect{w} )$ is minimized by using Lagrange multipliers $\lambda_i \geq 0$ for all $i = 1 \dots \; N$ and a Lagrangian of the form
\begin{displaymath}
  L(\vect{w}, b, \vect{\lambda}) = \frac{1}{2} ||\vect{w}||^2 - \sum_{i=1}^{N} \lambda_i (\: c_i (\langle \vect{w}, \vect{d} \rangle + b) - 1)) 
\end{displaymath}

\noindent
The Lagrangian $L$ is minimized with respect to the primal variables $\vect{w}$ and $b$, and maximized with respect to the dual variables $\lambda_i$. The derivatives of $L$ with respect to the primal variables must vanish and therefore, $\: \sum_{i=1}^{N} \lambda_i \: c_i = 0 \;$ and $\; \vect{w} = \sum_{i=1}^{N} \lambda_i \: c_i \: \vect{d_i} \;$.
The solution vector is a subset of the training patterns, namely those descriptor vectors with non-zero $\: \lambda_i \:$, called Support Vectors. All the remaining training samples $(\vect{d_i}, \: c_i)$ are irrelevant: their constraint $c_i (\langle \vect{w}, \vect{d_i} \rangle + b) \geq 1$ need not be considered, and they do not appear in calculating $\vect{w} = \sum_{i=1}^{N} \lambda_i \: c_i \: \vect{d_i}$. The desired hyperplane is completely determined by the descriptor vectors closest to it, and not dependent on the other training instances. 

\noindent
Eliminating the primal variables $\vect{w}$ and $b$ from the Lagrangian, we have the new optimization problem:
\begin{displaymath}
  \mathrm{minimize} \; \; \vect{W}(\vect{\lambda}) =  \sum_{i=1}^{N} \lambda_i - \frac{1}{2} \sum_{i,j=1}^{N} \lambda_i \: \lambda_j \: c_i \: c_j \: \langle \vect{d}, \vect{d_i} \rangle
\end{displaymath}
\hspace{2.5em} subject to $\lambda_i \geq 0$ for all $i = 1 \dots \; N \;$ and $\; \sum_{i=1}^{N} \lambda_i \: c_i = 0$

\smallskip \noindent
Once all $\lambda_i$ for $i = 1 \dots \; N \;$ are determined, the bias term $b$ is calculated. We construct a set $S$ of support vectors of size $s$ from descriptor vectors $\vect{d_i}$ with $0 \: < \: \lambda_i \: < \: 1$. This set of support vectors has equal number of descriptor vectors for $c_i \: = \: -1$ and for $c_i \: = \: +1$. The bias term is
\begin{displaymath}
b \; = \; - \frac{1}{s} \sum_{\vect{d} \in S} \sum_{j=1}^{N} \lambda_i \: c_j \: \langle \vect{d}, \vect{d_j} \rangle
\end{displaymath}

\smallskip \noindent
The classifier can thus be expressed as $f(\vect{d}) = \mathrm{sgn}(\sum_{i=1}^{N} \: \lambda_i \: c_i \langle \vect{d}, \vect{d_i} \rangle + b) \;$ for a descriptor vector $\vect{d}$. In our case, the descriptors are not linearly separable; therefore, they are mapped into a high-dimensional feature space using a mapping function or kernel. The hyperplane is computed in the high-dimensional feature space. The kernel $k$ is evaluated on the descriptors $\vect{d}$ and $\vect{d_i}$ as $k(\vect{d}, \vect{d_i}) \; = \; \langle \vect{d}, \vect{d_i} \rangle $. 

\noindent
Several kernels were tested---linear, sigmoid, homogeneous polynomial, Gaussian Radial Basis, and exponential kernels. Based on the performance of different kernels, we selected a Gaussian Radial Basis kernel for validation:
\begin{displaymath}
k(\vect{d}, \vect{d_i}) \; = \; \frac{1}{m} \sum_{k=1}^{m} e ^{\frac{(d_k-d_{ik})^2}{2 \: \sigma^2}}
\end{displaymath}

\section{Classifier Performance}
The training and testing sets are generated from visually verifying road vectors projected onto oblique images for geographical areas that have accurate camera projection parameters, road vectors, DEM, and 3-D building models. The positive class consisted of over 800,000 samples created from 1876 image regions and sampled on 192 road segments. These road segments were visible and well aligned with underlying image regions. A majority of these positive samples were from areas that did not have high-rise buildings and foliage; therefore, the corresponding oblique images had few occlusions and shadows. The negative class had over 500,000 samples generated from 2954 negative image regions---these regions were sampled from areas corresponding to building facades, foliage, and offset road vectors as observed in oblique images. 2500 samples from each class were randomly selected for testing the classifier’s accuracy. The rest of the samples were used for training. Selection, training, and testing steps are repeated over 80 such random splits of the data for the estimation of classification rates.

\begin{figure}[t]
  \centering
  \includegraphics[width=\linewidth]{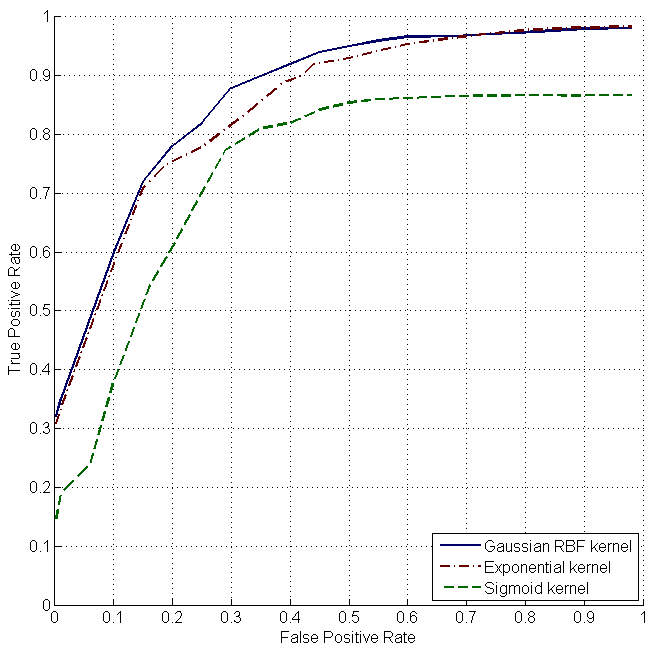}
  \caption{ROC curves for different SVM classifier kernels}
  \Description{ROC curves for different SVM classifier kernels}
  \label{Figure02}
\end{figure}

In our validation problem, the true positive rate, i.e., the rate of positive samples that are correctly classified, and the false positive rate, i.e. the rate of negative samples that are incorrectly classified as positive ones, can be used as measures of the validation classifier’s accuracy. We wish to achieve a high true positive rate while keeping the false positive rate low. The curve of the true positive rate versus the false positive rate is known as the Receiver Operating Characteristic (ROC) curve. ROC curves for different kernels are shown in Figure \ref{Figure02}. The area under a ROC curve is of significance; closer to 1.0 means better classifier performance. We found that the Gaussian Radial Basis Function (RBF) kernel has a better performance in classifying visibility of roads than other kernels such as linear, sigmoid, homogenous polynomial, and exponential radial basis kernel as can be observed in Figure \ref{Figure02}. The parameter $\sigma$ for the Gaussian kernel was set to 0.8. Performance of the SVM classifier with the Gaussian RBF kernel was numerically assessed by the following measures: 

\smallskip
\hspace{3em} sensitivity = $TP \; / \; ( TP + FN )$ 

\smallskip
\hspace{3em} specificity = $TN \; / \; ( FP + TN )$ 

\smallskip
\hspace{3em} accuracy = $( TP + TN ) \; / \; ( TP + FP + TN + FN )$

\smallskip
\hspace{-1em}where $TP$, $TN$, $FP$, and $FN$ respectively denote the number of descriptors classified as true positive, true negative, false positive, and false negative. Higher sensitivity values indicate better validation rates, whereas higher specificity values indicate better detection of erroneous or inconsistent data. The average sensitivity, specificity, and accuracy of our classifier are 89\%, 71\%, and 80\% respectively. Since pixel statistics are calculated locally, all image pixels need not be checked for validation or correction. Only pixel statistics along given vector data need to be examined for data validation. This makes our algorithm scalable, computationally efficient, and applicable to validating geospatial data using a variety of imagery. Analyzing vector data projected into an oblique image of 4000 x 3000 pixels taken in an urban region takes less than 400 milliseconds on a 3.4GHz. Intel Pentium 4, 2GB RAM machine. We have been able to scale this using a distributed cluster of machines because the oblique images and the image regions can be processed in parallel. 

One salient quality of SVMs is that the classifying hyperplane relies on only a small set of training examples or support vectors. However, finding these support vectors often requires an intelligent sampling of the training data for learning within large datasets in a reasonable amount of time. In non-linear SVM’s, time require for training the classifier increases cubically with the number of training examples. Finding support vectors and discarding descriptors that are unlikely to become support vectors is important for speeding up the training time of our SVM classifier and scaling it for large training datasets. This is a part of our ongoing effort towards improving this classifier’s performance during training.

\section{Results}
We have labeled a large repository of oblique images by combining camera projection parameters, digital elevation maps, 3-D building models, vector data, and landmark data. The labeling scheme, as seen in Figure \ref{Figure04}, accounts for occlusions, non-planarity of surface terrain, and foreshortening. Road labels are placed on visible road areas while building and landmark labels are placed on corresponding building areas in images, thus accounting for occlusions and clutter. Segments of roads that are occluded by buildings are marked by dashed lines. However, there are a few areas where solid roads markings appear on occluding buildings. Figure \ref{Figure04} shows a few such cases outlined in red. These errors are due to missing or outdated building models. Road segments in few other areas are not aligned with their corresponding pixel regions in the image. The misalignment between projected road segments and road pixels in images are due to inaccurate road vectors and road elevation (3-D terrain). We observed that errors or inaccuracies in vectors, DEM, building models, or camera projection parameters can contribute to misalignment of labeling in images.

\subsection{Data Validation}
Figure \ref{Figure05} shows the output of the SVM classifier; all visible road segments in the image were tested for data consistency. The segments in blue have descriptors that are consistent with well-aligned and visible-road pixel statistics. Vector, model, DEM, and map projection data are validated for all the blue road segments. The road segments marked in red are either occluded or misaligned with the underlying road pixel regions. The two areas that have missing or obsolete building models as highlighted in Figure \ref{Figure04} have been successfully identified by the classifier’s output. The road segment that is misaligned with its corresponding image region due to inaccurate vector data and altitude has been classified in red as well. 

Figure \ref{Figure06} shows the classifier detecting the misalignment between the projected road vector and the corresponding bridge in the image. Due to inaccurate DEM and the absence of the bridge’s 3-D model, the road vector snaps to the water surface. The projection of the road vector onto the image is, therefore, erroneous. Other areas where the road vectors do not line up with their image counterparts have been detected by the classifier. Regions where the road is occluded by foliage have been detected as well. The absence of exact altitude information for road nodes and Z-ordering of highways do not allow accounting for road occlusions at overpasses and tunnels. This is observed in Figure \ref{Figure07}, where the I-5 expressway is occluded by two overpasses before entering a tunnel. All occluded areas—under overpasses and tunnels—have been detected by our classifier in over 400,000 images. Please note that inconsistencies due to errors in altitude, such as in elevated freeways and ramps, cannot be easily detected in nadir or satellite imagery. Incorrect vector data, for example those updated by construction and public transportation projects, have been detected by the classifier’s output as well.

\section{Data Conflation}
Negative classification of a road segment could be due to one or a combination of inaccuracies in the following: road vector data, DEM or terrain model, 3-D building models. It could also be due to occluding structures such as foliage, tunnels, and overpasses. Inconsistencies and absence of information regarding any of the above geospatial entities will be detected by the classifier. A search for the correct projection of the road in the neighborhood of a segment that has been classified as negative or erroneous can be used for correcting the road segment. 

Figure \ref{Figure08} shows a segment of road that has been classified as negative. This road segment is sampled at regular intervals and the lines in green run perpendicular to the road segment at the sampled intervals. The classifier searches for the correct road segment along these lines in both directions of the incorrect road segment. The points encircled in black are the first points on the search lines that resulted in positive classification. Two sets of points are detected on both sides of the incorrect segment. The set of detected points above the incorrect segment, as can be noticed, corresponds to the flyover while the bottom set of detected points corresponds to the road underneath the raised flyover. Each set of points forms a smooth vector curve that exactly delineates the road in the image. 

The vector corresponding to the road on the bridge in Figure \ref{Figure09} is classified as inconsistent. This is due to the fact that the DEM sampled for the elevation of the road lies on the water surface as it does not include a 3-D model of the bridge. The classifier detects points, which are encircled in black, on the search lines perpendicular to the projected vector in the image. This set of points lies on one side of the incorrect vector and lines up with the road on the bridge. We have corrected over 40,000 road segments in the US by using our pixel-based conflation method.

Pixel statistics, which combine color, intensity gradients, texture, and steerable filter responses, encode appearance of visible roads and can be successfully used by our kernel classifier to robustly detect new roads in images. As observed in Figure \ref{Figure08} and \ref{Figure09}, the appearance of roads can vary considerably depending on lighting, camera properties, surrounding clutter, surface albedo, and reflectance. The redundancy built into the descriptors and the kernel used for mapping the descriptors into the feature space help in accounting for these variations. At the same time, the descriptors are simple and not computationally intensive.

\section{Related Work}

Our vector-validation method learns pixel statistics for determining consistency of geospatial data. Learning pixel statistics using a kernel classifier makes our algorithm robust, explainable, and widely applicable to different kinds of imagery. This approach need not be limited to oblique imagery; it can be applied to nadir or satellite imagery, or ground level imagery as well. There is no restriction on the camera angle of the images. Our validation algorithm can be successfully applied to data validation of top-down satellite or nadir imagery as well. We have selected oblique imagery primarily due to two reasons: first, oblique images capture structure, altitude, terrain, and 3-D attributes of the scene; and second, oblique images capture view-dependency and therefore, are able to register relative positions of geospatial structures through occlusions. 

The accurate alignment of imagery, maps, and vector data from disparate sources are usually based on various conflation methods \cite{ConflationMap1988, ConflatingOrthoimageryStreetMaps2004, AnnotatingSpatialDatasets2003, IntegrationVector1998}. Conflation methods use a set of control point pairs identified in the disparate data sources. Until now, vector data has been conflated with satellite imagery, where the control points have been determined manually and automatically using features such as straight roads and road intersections. Our paradigm for correction of vector data is markedly different from other conflation methods. We learn pixel statistics of visible roads and check if the statistics are consistent along the putative road segment. We use the learnt pixel statistics to detect inconsistencies in vector data, building models, camera parameters, and digital terrain data. The same statistics can be used to correct vector data if we assume that the inconsistency is solely due to incorrect vector data. Our algorithm does not make any assumption about the transformation between the data sources. For example, we do not assume that the transformation between the vector data and imagery is a simple translation or an affine transformation \cite{ConflationMap1988, AutomaticAlignmentRoadMaps2007}. In fact, the transformation between vector data and its projection in oblique imagery is non-linear, and we leverage that property to validate our vector data. 

Our algorithm does not use linear road features \cite{IntegrationVector1998, AutomaticAlignmentRoadMaps2007} or a few road intersection points \cite{ConflatingOrthoimageryStreetMaps2004} as control points; instead, it uses local pixel statistics that combine color, feature gradients, second-order derivatives of pixel intensities, and texture cues to validate and correct road pixels and road data. In contrast, notable conflation algorithms \cite{ConflationMap1988, ConflatingOrthoimageryStreetMaps2004, RoadGridExtraction1999} match control points, such as special point-features or line-features, derived from two disparate datasets. Many conflation algorithms also require manual selection of control points for registration of two disparate data sources which do not make them scalable. In some prior research, control points have also been automatically selected using localized processing of pixels in satellite imagery \cite{AnnotatingSpatialDatasets2003}.

Oblique imagery has not been used for conflation in any prior art (\cite{AutomaticRoadExtraction2003} and references therein). Conflation techniques were limited to nadir or satellite imagery until now. Prior research in conflation is not applicable to oblique imagery because the appearance of image features varies greatly in oblique imagery due to the perspective or oblique view. Roads that have distinct straight-line features or intersections in top-down views do not appear the same in oblique imagery due to the non-linear projection of the 3-D world
into the 2-D image and surrounding clutter. Furthermore, the presence of regular texture, mostly straight-line features, on building facades makes it impossible to do conflation using corners, edges, or straight-line image-based features. For example, conflation schemes that use straight line features often snap the road to the shadow of the bridge on the water surface as seen in Figure \ref{Figure06} and \ref{Figure09}. 

Our algorithm finds inconsistencies and inaccuracies in 3-D models and DEM (terrain) as well. Unlike all prior research, we do not limit our scope to detecting inconsistencies due to incorrect vector data alone. Image segmentation is not done at any stage of our algorithm. Image segmentation is expensive and does not work well for high-resolution imagery---oblique or otherwise---due to the presence of clutter and variability in image regions such as presence of foliage or traffic. Our algorithm does not perform image segmentation to identify spatial entities or objects \cite{AnnotatingSpatialDatasets2003, RoadNetworkExtraction2007}.

Structure-from-Motion (SfM) or stereo algorithms compute the 3-D model or structure of the scene. Unlike SfM and stereo, our algorithm does not model roads, building models, or terrain from multiple images to determine their consistency. Instead, it uses a single image to determine data consistency by using a classifier. This makes it especially useful for data validation using new user-contributed images. We can also apply the same algorithm to multiple or a sequence of images, where the validation scheme can be further strengthened by combining inferences from multiple images.

\section{Conclusion}
Oblique images are a rich source of geospatial information. They also capture the view dependency of geospatial entities; relative positions, heights, structures, and extents of geospatial entities are better observed in oblique images when compared to satellite or nadir imagery. We believe that oblique views of the world will play an increasingly significant role in many mapping applications. This paper presents the novel use of oblique images in validating geospatial data.

\begin{figure}[t]
  \centering
  \includegraphics[width=\linewidth]{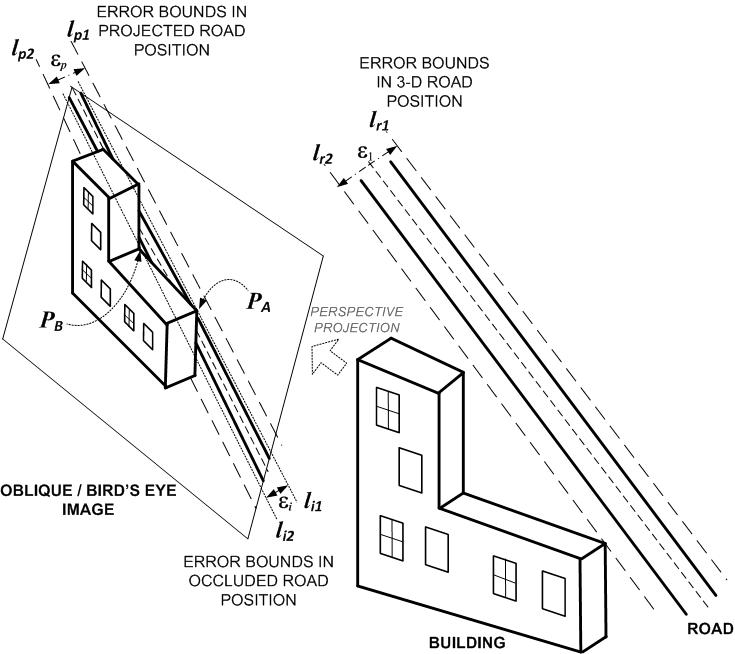}
  \caption{A road and building as viewed in the 3-D world in an oblique image along with the error bounds}
  \Description{A road and building as viewed in the 3-D world in an oblique image along with the error bounds}
  \label{Figure03}
\end{figure}

\begin{figure*}
  \includegraphics[width=0.82\textwidth]{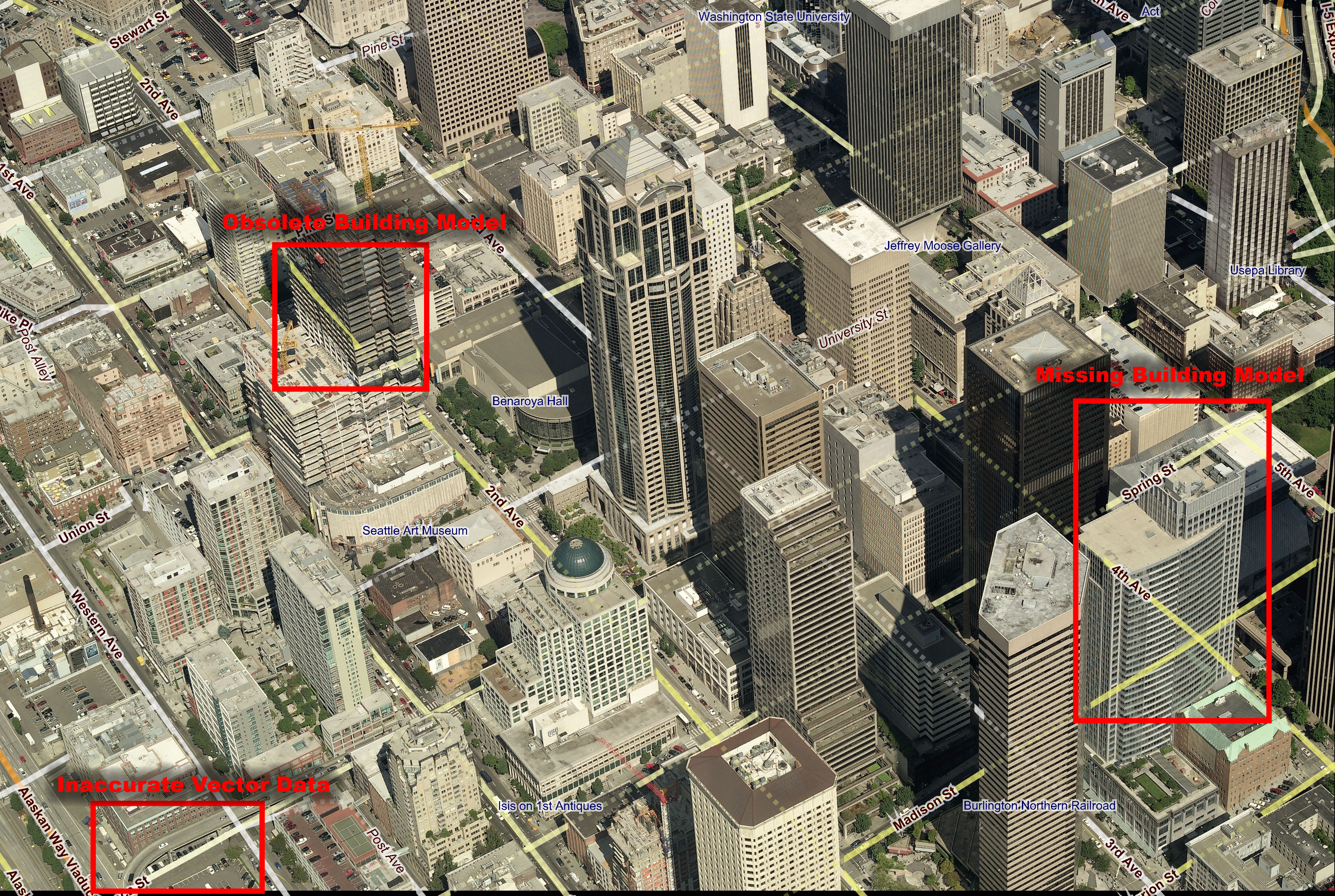}
  \caption{Vector data is projected onto an oblique image by accounting for terrain and occlusions.}
  \label{Figure04}
  \vspace{0.8cm}
\end{figure*}

\begin{figure*}
  \captionsetup{width=.82\linewidth}
  \includegraphics[width=0.82\textwidth]{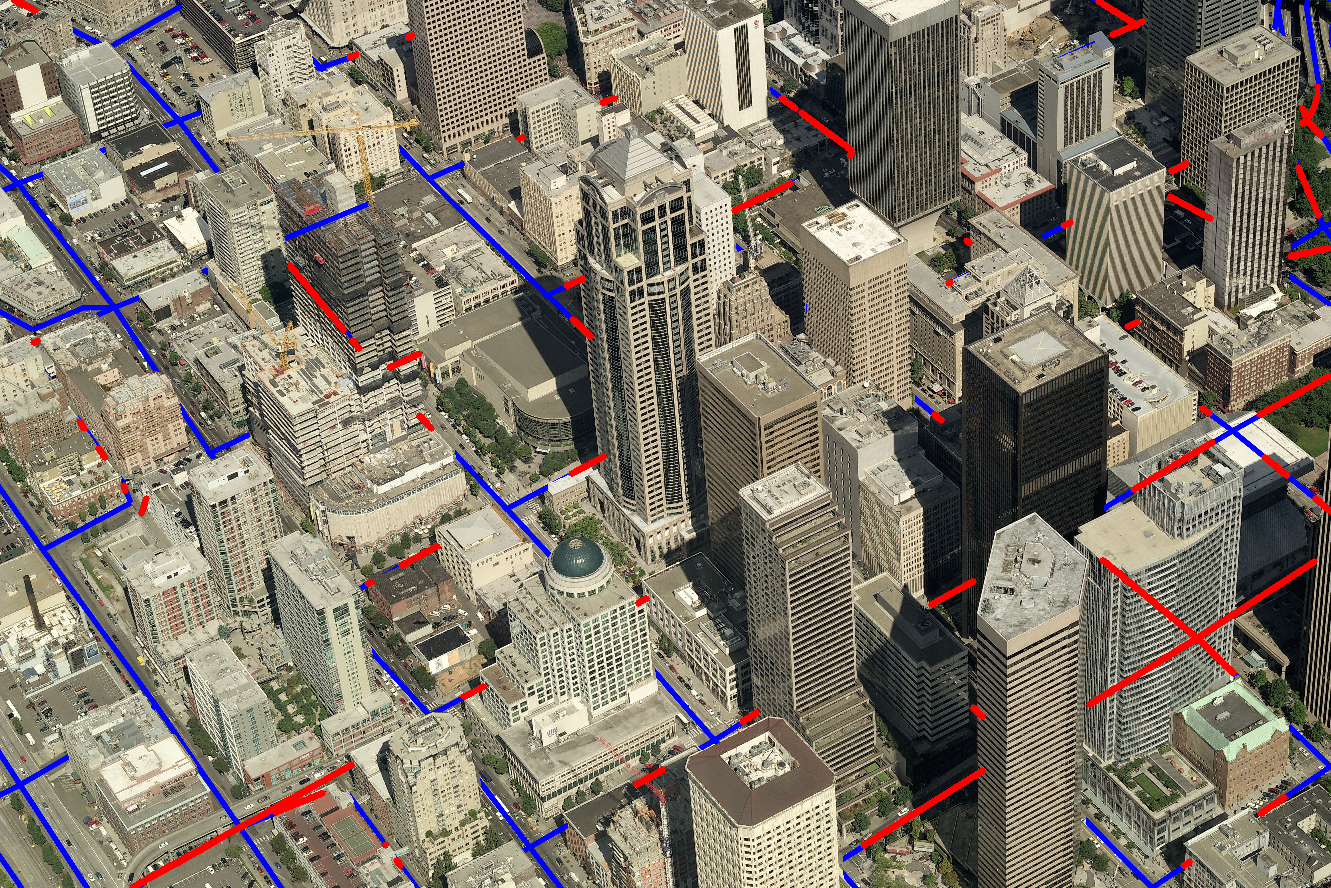}
  \caption{Road segments consistent with corresponding pixels in the oblique image are classified in blue while inconsistent segments are in red. Inconsistencies highlighted in Figure \ref{Figure04} have been detected.}
    \label{Figure05}
\end{figure*}

Oblique images capture the subtle spatial interdependency between different geospatial entities that are in close proximity. Perspective and foreshortening effects, occlusions, and relative delineation of image features arise from this interdependency. To illustrate the occlusion aspect of this interdependency, consider a simple example of a road being occluded by a building as illustrated in Figure \ref{Figure03}. A road in 3-D space has error bounds in its position $\vect{l_{r1}}\;$ and $\vect{l_{r2}}\;$ with respect to the centerline. The uncertainty in its position is $\vect{\varepsilon_1}$. As observed in the oblique image, the projection of the road in the image has uncertainty $\vect{\varepsilon_p}$. Note that this projection is a non-linear (perspective) and takes into account the altitude, in addition to latitude and longitude, of points on the road. The error bounds undergo a transform and are projected on the image as $\vect{l_{p1}}$ and $\vect{l_{p2}}$. The road is occluded by another geospatial entity---a building in this illustration---in the oblique view. If the projected road’s pixel statistics are used to determine this interdependency between the two geospatial entities, the fact that the road is being occluded by the building can be used to reduce the error bounds of the road. Note that this reduction in error bounds does not take into account any image features, which can further reduce the error bounds. The projected error bounds are limited to $\vect{l_{i1}}$ and $\vect{l_{i2}}$ at points $\vect{P_A}$ and $\vect{P_B}$. The uncertainty is reduced to $\vect{\varepsilon_i}$ as $\vect{\varepsilon_i}$ < $\vect{\varepsilon_p}$. That is to say, given the view-dependency of 3-D geospatial entities in oblique images, inconsistencies in observed data can be detected and corrected at a much finer level. 

We have shown that richness of geospatial information in oblique images can be leveraged for validating and correcting geospatial-vector data by analyzing pixel statistics of regions that correspond to geospatial entities. Image regions belonging to different entities have characteristic patterns and statistics, which help in discriminating one region from another, and therefore, one entity from another. However, analyzing pixel statistics for inferring consistency of an observed geospatial entity requires understanding and modeling the variability in appearance of that entity in different images. Our SVM classifier accounts for this variability and makes it possible to discriminate between different entities while enforcing consistency in the appearance of a given entity.

The advantage of using SVM is that the discriminating hyperplane and therefore, the classification depend on a small set of training descriptors, or the support vectors. In our application, these support vectors are those descriptors that are responsible for separating two classes of geospatial entities. These support vectors encode the characteristic appearance of a given entity, which makes them useful for validation of that entity using images. The small set of support vectors makes our approach computationally efficient. We analyze consistency of a geospatial entity in an oblique image by combining color, pixel-intensity gradients, texture, and steerable filter responses---all of which contribute to the appearance of the entity. Consistency of the entity in an image or multiple images validates vector and elevation attributes of the entity, DEM and 3-D model data. The 3-D model data may also include occluded and occluding models in the neighborhood of the entity.

\begin{figure*}
\captionsetup{width=0.8\linewidth}
  \includegraphics[width=0.8\textwidth]{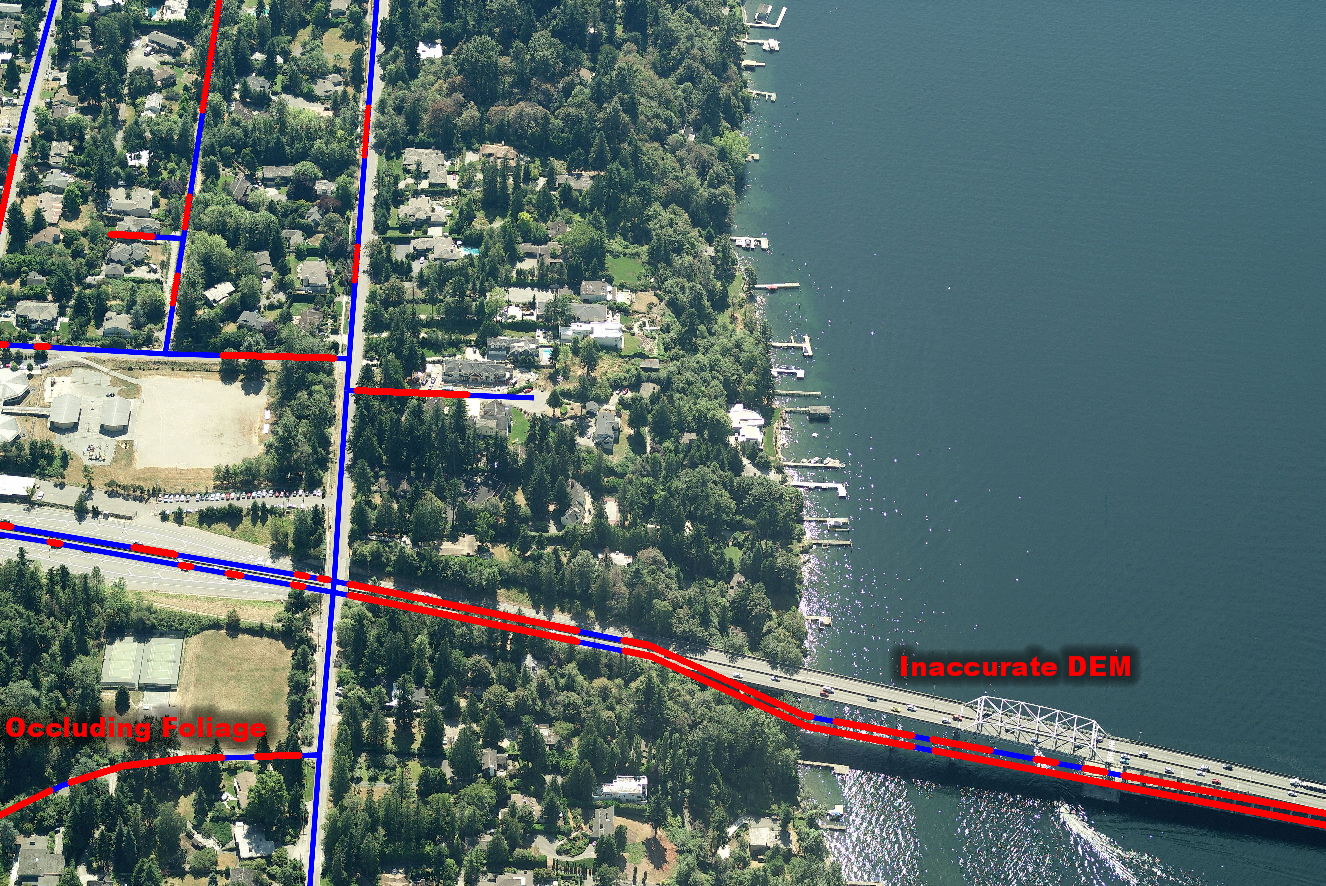}
  \caption{Road segments that are occluded by foliage or misaligned with their corresponding image areas due to inaccurate DEM are classified in red while consistent road segments are marked in blue.}
  \label{Figure06}
  \vspace{0.8cm}
\end{figure*}

\begin{figure*}
  \captionsetup{width=0.8\linewidth}
  \includegraphics[width=0.8\textwidth]{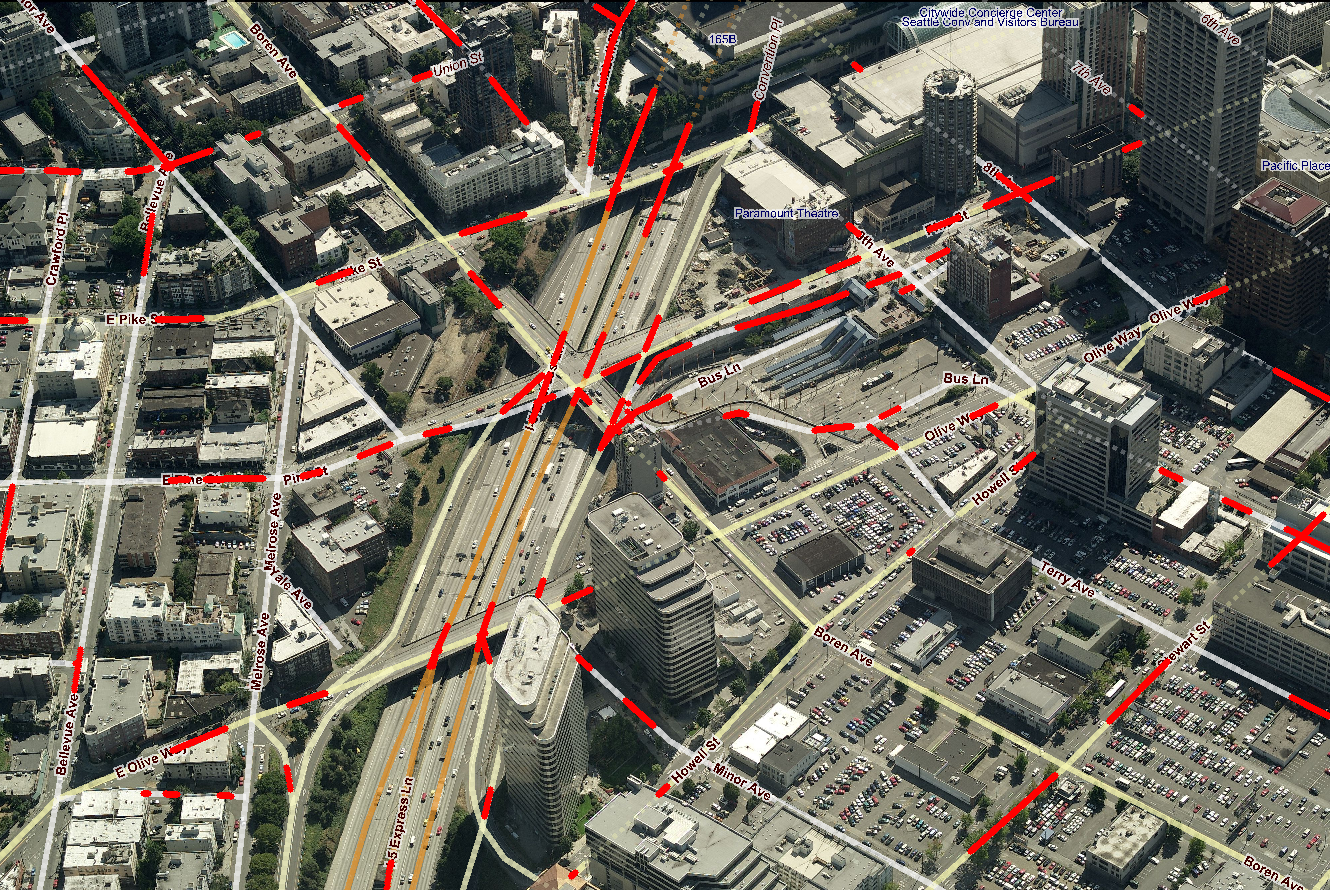}
  \caption{Vector data that do not exactly line up with their counterparts in the image due to occluding tunnels, absent Z-ordering, and inaccuracies are marked in red.}
  \label{Figure07}
\end{figure*}

\begin{figure*}
  \captionsetup{width=0.8\linewidth}
  \includegraphics[width=0.8\textwidth]{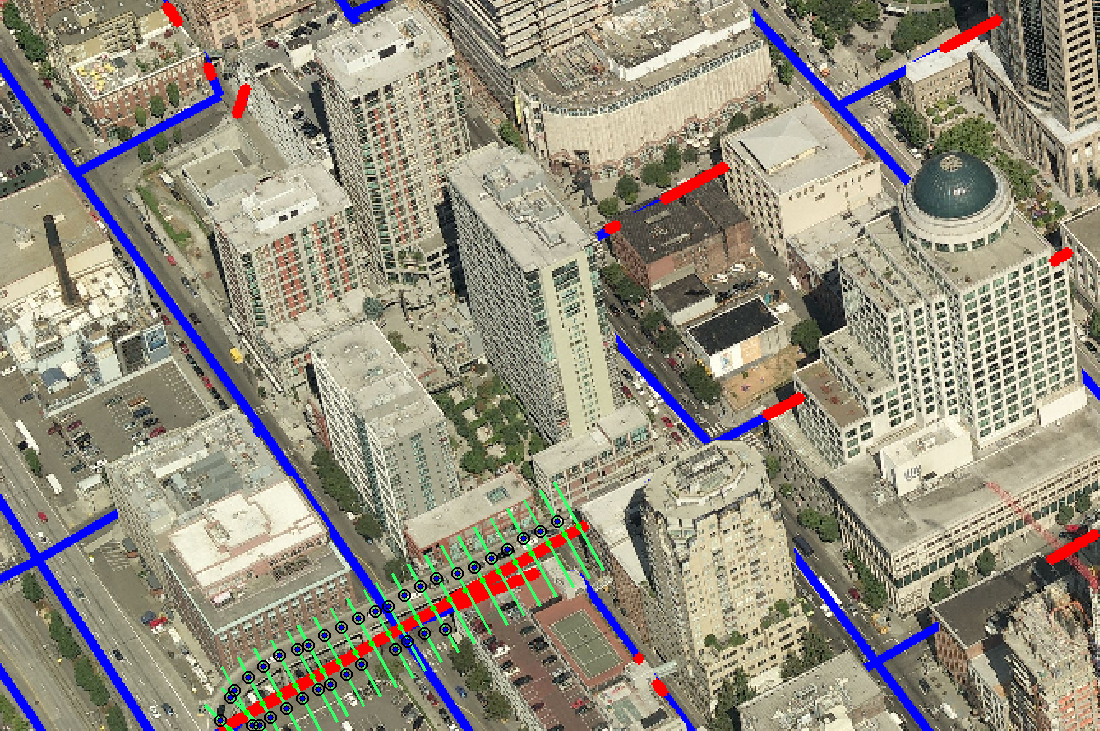}
  \caption{The classifier is applied along the green lines that are perpendicular to each inconsistent road segment. The encircled points in black delineate the new road vectors.}
  \label{Figure08}
  \vspace{0.8cm}
\end{figure*}

\begin{figure*}
  \captionsetup{width=0.80\linewidth}
  \includegraphics[width=0.8\textwidth]{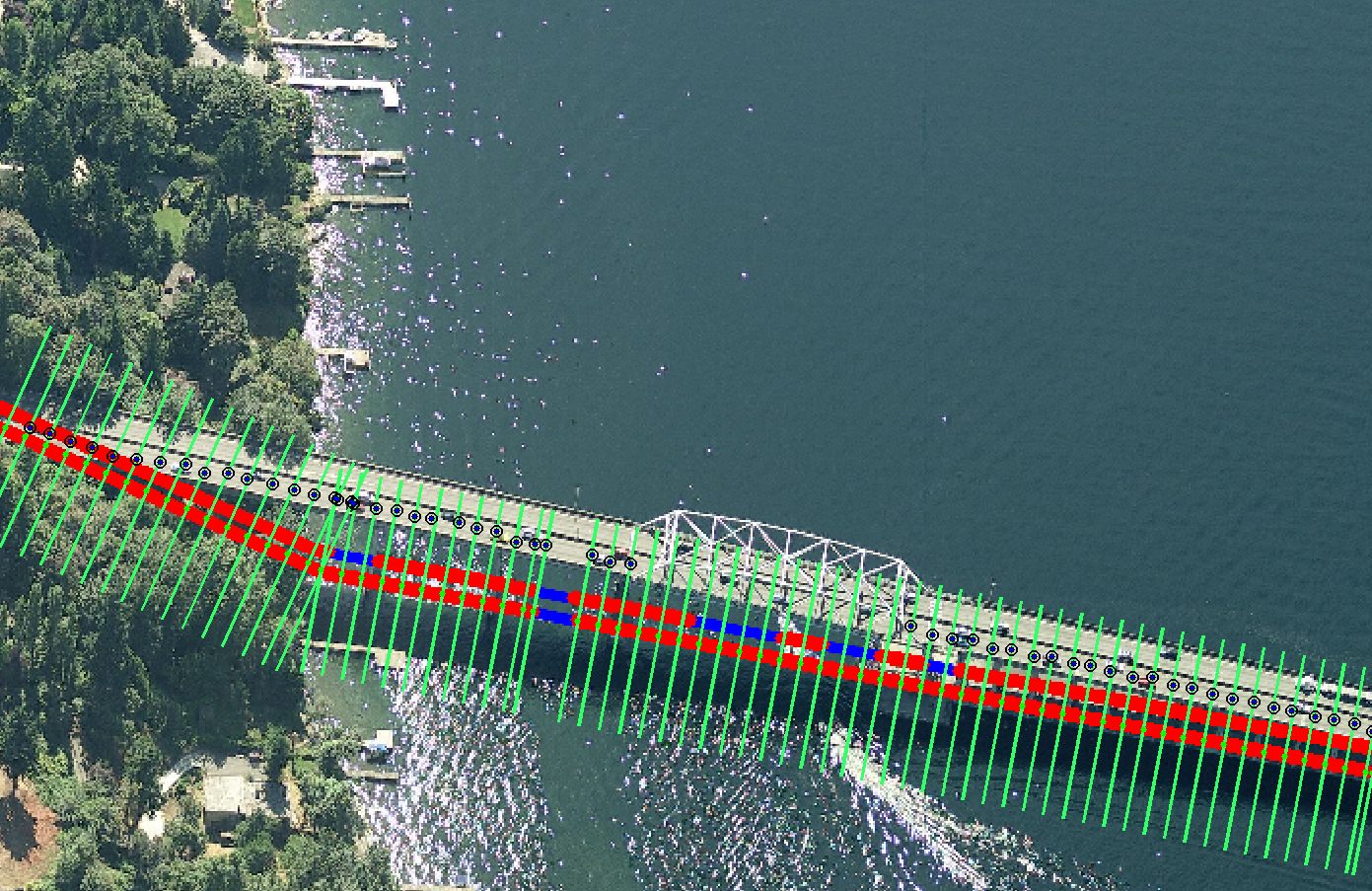}
  \caption{A search for the first point along each green line where the classifier responds positive yields the encircled point in black. These points delineate the corrector vector of the road on the bridge.}
  \label{Figure09}
\end{figure*}

A few unique challenges posed by oblique views include occlusions due to foliage and shadows. Figure \ref{Figure06} shows that road segments that are occluded by foliage have been classified as negative. The same kernel classifier is trained with occlusion data due to foliage, with hand-labeled image regions from images such as the one in Figure \ref{Figure03}. We do not have any 3-D data that include scanned foliage and ever-changing vegetation. A road and building as viewed in the 3-D world in an oblique image along with the error bounds can be used for detecting vegetation. Foliage has characteristic color and texture that can be encoded into our descriptor. Shadows pose a difficult problem because the pixel regions of shadows can be close to low-albedo regions; additionally, pixel-intensity gradients within shadow regions tend to be weak. 

We have presented an approach to validate vector data using descriptors derived from oblique images. These descriptors combine multiple cues from color, pixel-intensity gradients, texture, and steerable filter responses. A SVM classifier with a Gaussian-RBF kernel learns the descriptors that are consistent with observed geospatial data. It then validates geospatial data using new oblique images and detects those with inconsistent descriptors. 

Our oblique-view, vector-validation approach can be improved and extended further. It can be applied to validation of data using satellite and user-contributes images. Inconsistencies due to different sources of data can be identified and categorized using the same classifier---such as inconsistencies due to inaccurate DEM, missing building models, misaligned vector data, inaccurate map projection parameters, occlusions due to foliage, and so on. It can be used for validating building facades and therefore, 3-D models, boundaries of water bodies and vegetation, and other visible geospatial attributes. Collecting pertinent training data is necessary for these extensions. 

Finally, we believe that a wide field-of-view image taken at an oblique angle is a rich source of geospatial information. Oblique viewpoints capture the 3-D nature and view-dependency of geospatial data. Validating geospatial data by enforcing consistency between geospatial entities and their appearance in oblique images is a significant step towards extracting information from this unique data source.

\printbibliography

\end{document}